\documentclass[]{spie}  
\usepackage{algpseudocode}
\usepackage{algorithm}
\usepackage[]{graphicx}
\usepackage{amsmath}
\usepackage{subcaption}
\usepackage{caption}
\usepackage{fancyhdr}

\fancypagestyle{title}{%
    \setlength{\headheight}{22pt}%
    \fancyhf{}
    \fancyhead[R]{\begin{tabular}{@{}l} Distribution Statement A: Approved for 		\\ public release. Distribution is unlimited. \end{tabular}}
}%

\usepackage[headsep=0.5cm]{geometry}

\title{Investigation of Initialization Strategies for the Multiple Instance Adaptive Cosine Estimator} 

\author{James Bocinsky, Connor McCurley, Daniel Shats, and Alina Zare
\skiplinehalf
Department of Electrical \& Computer Engineering, University of Florida
}

\authorinfo{Further author information: (Send correspondence to James Bocinsky and Alina Zare)\\James Bocinsky: E-mail: jbocinsky1@gmail.com\\  Alina Zare: E-mail: azare@ece.ufl.edu}

\begin{document} 
  \maketitle 
  \thispagestyle{title}

\begin{abstract}
Sensors which use electromagnetic induction (EMI) to excite a response in conducting bodies have long been investigated for subsurface explosive hazard detection. In particular, EMI sensors have been used to discriminate between different types of objects, and to detect objects with low metal content. One successful, previously investigated approach is the Multiple Instance Adaptive Cosine Estimator (MI-ACE). In this paper, a number of new initialization techniques for MI-ACE are proposed and evaluated using their respective performance and speed. The cross validated learned signatures, as well as learned background statistics, are used with Adaptive Cosine Estimator (ACE) to generate confidence maps, which are clustered into alarms. Alarms are scored against a ground truth and the initialization approaches are compared.  
\end{abstract}



\section{INTRODUCTION}

Many machine learning approaches use individual samples with their corresponding classification label to learn a representative of each class for classification. For some applications, such as explosive hazard detection\cite{zare2015multiple} \cite{karem2015fuzzy} \cite{alvey2017fourier} \cite{cook2016buried} or drug activity prediction\cite{dietterich1997solving}, this is unfeasible due to the ambiguity of the data. The data does not present itself in a way that can be labeled at the sample level. A machine learning framework known as multiple instance learning (MIL) was formalized by Dietterich et al. to handle applications with ambiguously labeled data \cite{dietterich1997solving}. In this framework, instance level labels are not available, so the data is grouped into bags, with bag level labels. Although there are other variations to MIL, the original MIL framework assumes that a ``positive'' bag corresponds to a bag that has at least one instance corresponding to a target class of interest. A ``negative'' bag corresponds to a bag that contains only instances that belong to the background, non-target class. A typical process of using MIL is to determine the instances and their salient features from the positive bags that help differentiate the detection of unknown instances.

In the context of explosive hazard detection, one reason MIL is often used is because the area of the explosive hazard's EMI response is unknown. It is true that the location of a target is known, but it is unknown how the magnitude of the response will reduce over space, or how much spatial area the target response can be detected. This invokes one reason why MIL fits this application; it is simple to label a spatial region as containing a response from a explosive hazard as a ``positive'' bag, and other regions where no explosive hazards exist as ``negative'' bags. Labeling individual responses, known as instances in MIL, along a physical sweep of a handheld sensor is nearly impossible. Additionally, another benefit of using MIL for explosive hazard detection is to allow the algorithm to learn from the data what is a good representative of an explosive hazard. Laboratory measurements can be measured or physics based models like the Discrete Spectrum of Relaxation Frequencies\cite{wei2011landmine} can be generated to model an explosive hazard response. However, it is often seen that the response of the electromagnetic induction (EMI) sensors used for explosive hazard detection act differently when an explosive hazard is buried than in the lab or physics based models. So although these models may be accurate in the lab or in ideal physical conditions, various soil responses often make these models inaccurate to real world responses. More so, MIL can learn a representative that is normalized to the soil and model the local soil properties. This way, the MIL algorithm can learn a representative that maximizes detection in the soil being investigated. 

The Multiple Instance Adaptive Cosine Estimator (MI-ACE) algorithm is a MIL algorithm that makes use of these properties, and even more so focuses on maximizing target detection using the Adaptive Cosine Estimator (ACE) detection statistic. This means that the algorithm does not only determine a signature that is most like a target signature and unlike the background data, but as well learns a target signature that maximizes the ACE detection statistic. The ACE statistic has nice properties for explosive hazard detection. ACE applies a transformation to every test sample with respect to the background data, known as data whitening.\cite{mayer2003object}\cite{kraut2001adaptive}\cite{wu2016two} This is done by subtracting the background mean and multiplying by the inverse covariance matrix of the background. This results in a zero mean and unit variance data point with respect to the background. ACE also normalizes the similarity by the magnitude of both the target signature and the test sample. This means that regardless of magnitude, the relative feature vector values or shape of the test sample is responsible for the similarity measure. This is highly beneficial for detecting low metal explosive hazards where the magnitude response may be low. Furthermore, explosive hazards buried at different depths often have similar response shape, but at different magnitudes. Since the shape of the response is used for detection, these explosive hazards can still be found. With these properties, it can be seen that the MI-ACE algorithm has many desired aspects of an ideal algorithm for explosive hazard detection.

One aspect of MI-ACE that is crucial to the performance of the algorithm is the initialization procedure. The optimization part of the algorithm is dependent on there being a reasonably well initialized target signature. If not, the algorithm may converge to a poor representative. The principle initialization technique outlined in the original MI-ACE publication\cite{zare2018discriminative} is to initialize the instance from the positive bag that maximizes the MI-ACE objective function. This requires searching through all of the positive instances, finding each positive instance's positive bag representative, computing the ACE similarity to each of the background samples, and finally computing the objective function. This is computationally expensive with a complexity of $O((N^{+})^{2} + N^{+}N^{-})$, where $N^{+}$ and $N^{-}$ are the total number of positive and negative instances respectively.

In this paper, three new initialization techniques are investigated to improve performance and reduce the computational cost of the initialization process. Clustering techniques are investigated to reduce the number of samples that must be searched for initialization. As well, a new statistic, the multiple instance cluster rank is proposed to reduce the computation complexity of initializing a target representative. The performance of these techniques with the initialized and optimized signatures, for three data sets, and two sensors are shown along with a computational cost analysis.

\section{METHODS}

\subsection{Multiple Instance Adaptive Cosine Estimator}
The Multiple Instance Adaptive Cosine Estimator (MI-ACE) was originally proposed to solve common problems of spatial inaccuracy in training data in target detection applications \cite{zare2018discriminative}. In order to use a detection metric like ACE, a target signature must be known prior to performing detection. Techniques to estimate target representatives can be measured in a laboratory setting, but are often unrealistic and not representative of a target in various conditions and environments. Alternatively, a target representation may be extracted directly from the data itself. Often times when this is done, the extracted representation does not contain meaningful features to differentiate it from the background and may not provide the desired performance. Lastly, it is often times difficult or even impossible to extract a target representation from a dataset. For the explosive hazard detection problem, the boundaries of an explosive hazard's response within a physical sweep of the sensor are challenging to obtain, and thus determining where to extract a target representation is nonviable. The MI-ACE algorithm addresses these problems and is able to learn a target signature that is optimal for the ACE detection metric.

\subsubsection{Method}

The MI-ACE algorithm\cite{zare2018discriminative} follows the multiple instance learning framework where the labels of the data are at the bag level. With this, let $\mathbf{X} = [\mathbf{x_{1}}, ..., \mathbf{x_{N}}]$ be training data with each sample, $\mathbf{x_{i}}$ being a vector with dimensionality $d$. The data is grouped into $K$ bags $\mathbf{B} = \{\mathbf{B_{1}},..., \mathbf{B_{K}}\}$ with labels, $L = \{L_{1}, ..., L_{K}\}$, where $L_{j} \in \{0, 1\}$. A bag is considered positive, $\mathbf{B_{j}^{+}}$, with label, $L_{j} = 1$, when there exists at least one instance in bag $j$ that is from the target class. Additionally, a bag is considered negative, $\mathbf{B_{j}^{-}}$, with label $L_{j} = 0$, if all instances in bag $j$ are from the background class. The number of instances in both positive and negative bags does not need to be fixed.

With this formulation, the goal of MI-ACE is to estimate a target signature, $\mathbf{s}$, that maximizes the detection statistic of the target instances in the positive bags while minimizing the detection statistic of all negative instances. This is accomplished by maximizing the objective shown in eq. \eqref{MIACEobjfun},

\begin{equation}
    \arg\max_{\mathbf{s}} \frac{1}{N^{+}} \sum_{j:L_{j} = 1} D_{ACE}(\mathbf{x_{j}^{*}}, \mathbf{s}) - 
    \frac{1}{N^{-}} \sum_{j:L_{j} = 0} \frac{1}{N_{j}^{-}} \sum_{\mathbf{x_{i}} \in B_{j}^{-}} D_{ACE}(\mathbf{x_{i}}, \mathbf{s}) ,
    \label{MIACEobjfun}
\end{equation}
where $N^{+}$ is the number of positive bags, $N^{-}$ is the number of negative bags, and $N_{j}^{-}$ is the number of instances in negative bag $j$. $\mathbf{x_{j}^{*}}$ is the positive instance selected from bag $j$ that is most like the target signature, $\mathbf{s}$, known as the bag representative,

\begin{equation}
    \mathbf{x_{j}^{*}} = \arg\max_{\mathbf{x_{i}} \in \mathbf{B_{j}^{+}}} D(\mathbf{x_{i}}, \mathbf{s}) \text{.}
    \label{MIACEBagRep}
\end{equation}

The detection statistic, ACE, shown in eq. \eqref{ACEDet}, is the projection of a test sample, $\mathbf{x}$, onto a known target signature, $\mathbf{s}$, in a whitened coordinate space. Again, the whitening is done using the background covariance, $\boldsymbol{\Sigma_{b}^{-1}}$, and background mean, $\boldsymbol{\mu_{b}}$, to transform the data to have zero mean and a uniform, unit variance with respect to the background. ACE is normalized by not only the target signature, $\mathbf{s}$, but also the whitened test sample, $\mathbf{x}$, as well. With this, the magnitude of the test sample will not affect the statistic, and only the shape of the feature vector contributes to the statistic.

\begin{equation}
    D_{ACE}(\mathbf{x}, \mathbf{s}) = \frac{\mathbf{s^{T} \Sigma_{b}^{-1}} (\mathbf{x} - \boldsymbol{\mu_{b}})} {\sqrt{\mathbf{s^{T} \Sigma_{b}^{-1} s}} \sqrt{(\mathbf{x} - \boldsymbol{\mu_{b}})^{\mathbf{T}} \mathbf{\Sigma_{b}^{-1}} (\mathbf{x} - \boldsymbol{\mu_{b}})}}
    \label{ACEDet}
\end{equation}

To estimate the target signature, the objective function in eq. \eqref{MIACEobjfun} is maximized. To accomplish this, the algorithm is broken up in to two primary steps, initializing a target signature, and then optimizing that signature using a single instance from each positive bag, also known as the bag representative, $\mathbf{x_{j}^{*}}$. The original initialization method computes the objective function for all of the positive instances and whichever instance provides the largest objective function becomes the initialized target signature. Although this instance may provide the highest objective function value, it may not be optimal for all of the positive instances within the data. So considering this, optimization is done using the update equation shown in eq. \eqref{MIACEOpt}. Here $\mathbf{\hat{\hat{s}}}$, $\mathbf{\hat{\hat{x_{j}}}^{*}}$, and $\mathbf{\hat{\hat{x}}_{i}}$ are the whitened signature, whitened bag representative, and whitened negative instance respectively.

\begin{equation}
    \mathbf{\hat{\hat{s}}} = \frac{\mathbf{t}}{\vert\vert \mathbf{t} \vert\vert}
    \quad \text{where} \quad
    \mathbf{t} = \frac{1}{N^{+}} \sum_{j:L_{j} = 1} \mathbf{\hat{\hat{x}}_{j}^{*}} - 
    \frac{1}{N^{-}} \sum_{j:L_{j} = 0} \frac{1}{N_{j}^{-}} \sum_{\mathbf{x_{i}} \in B_{j}^{-}} \mathbf{\hat{\hat{x}}_{i}}
    \label{MIACEOpt}
\end{equation}

To optimize the initialized signature, the signature, $\mathbf{\hat{\hat{s}}}$, is iteratively updated using eq. \eqref{MIACEOpt}. In each iteration, the current bag representatives, $\mathbf{x_{j}^{*}}$, are determined given the current estimated target signature. The bag representatives are averaged and then the average of the background samples is subtracted away. The average background will not change from iteration to iteration so this term can be precomputed. Finally the target signature is normalized and the updated target signature has been computed.

\subsection{Alternative Initialization Approaches}\label{SectionApproaches}

Additional initialization approaches using clustering methods have been investigated to determine if either performance or run time can be improved for MI-ACE. The original initialization approach is to initialize the instance from a positive bag that maximizes the objective function. This requires the MI-ACE algorithm to search through all of the positive instances and compute the objective function. This is a computational expensive process, $O(N^{+}(N^{+}+N^{-}))$, where $N^{+}$ and $N^{-}$ are the total number of positive and negative instances respectively. Alternative initialization approaches using clustering are explored to reduce computation time by using the cluster centers as target candidates instead of every positive instance. With this, the algorithm does not need to search through as many candidate points to initialize a target signature. Additionally, the initialization technique will learn a representation of the target signature that is representative of a subspace of the data, instead of initializing a single instance that may or may not represent a greater region of the target class.

\subsubsection{K-Means}

The first approach, referred to as K-Means, uses the K-Means clustering algorithm\cite{macqueen1967some} to group all of the data, regardless of bag structure, into $K$ clusters. Then, this initialization approach picks the cluster center that maximizes the MI-ACE objective function as the initialized target signature. This computation complexity is $O(Ki(N^{+}+N^{-}) + K(N^{+} + N^{-}))$, where $K$ is the number of clusters, $N^{+}$ and $N^{-}$ are the number of positive and negative instances, respectively, and $i$ is the number of iterations until K-Means converges. The first term corresponds to K-Means clustering, and the second term corresponds to determining the cluster centers that maximize the objective function. As long as the number of clusters, $K$, and the number of iterations $i$, remains small, the K-Means approach will have less of a computational cost than the original initialization method. This way the algorithm only needs to search through $K$ candidates instead of $N^{+}$ candidates to initialize a target signature.

\subsubsection{Ranked K-Means}

The second approach, referred to as Ranked K-Means, uses the K-Means clustering algorithm\cite{macqueen1967some} to create $K$ clusters, regardless of bag structure. Instead of using the original objective function to score the cluster centers, a new multiple instance cluster rank is proposed to further reduce the computational cost. The multiple instance cluster rank of the $k^{th}$ cluster, is the sum of the proportions of the elements in cluster $k$. The three terms of the rank are the proportion of positive bags that have an instance in cluster $k$, the proportion of instances in cluster $k$ that came from a positive bag, and the proportion of instances in cluster $k$ that came from a negative bag. This is formally defined below in \eqref{MIClusterRank}

\begin{equation}
	Rank_{MIC}(k) = \frac{w^{{B^{+}}} \bigg(\frac{N^{B^{+}_{k}}}{N^{B^{+}}}\bigg)
 + w^{+} \bigg(\frac{N^{+}_{k}}{N^{+}}\bigg)
 - w^{-} \bigg(\frac{N^{-}_{k}}{N^{-}}\bigg) + w^{-}}
 {w^{B^{+}} + w^{+} + w^{-}} ,
	\label{MIClusterRank}
\end{equation}
where $N^{B^{+}}$, $N^{+}$, $N^{-}$ are the total number of positive bags, positive instances, and negative instances, respectively. $N^{B^{+}}_{k}$, $N^{+}_{k}$, $N^{-}_{k}$ are the number of positive bags that have at least one instance in cluster $k$, the number of positive instances in cluster $k$, and the number of negative instances in cluster $k$, respectively. Finally, the weights $w^{B^{+}}$, $w^{+}$, and $w^{-}$ are positive hyperparameter weights that are set based on the distribution of instances in the constructed positive bags. If it is believed that the positive bags contain a majority of positive instances, then the first two weights should be higher than the last weight. Furthermore, if the positive bags contain a majority of negative instances, the last weight should dominate to require a minimal number of negative instances belonging to the cluster center that is initialized. The equation adds $w^{-}$ and is divided by the sum of the weights to force the rank to be from $[0,1]$.

The computation complexity for this technique is $O(Ki(N^{+}+N^{-}) + K)$, where $K$ is the number of clusters, $N^{+}$ and $N^{-}$ are the number of positive and negative instances, respectively, and $i$ is the number of iterations until K-Means converges. In this initialization technique, the first term, corresponding to K-Means, dominates the complexity. The second term corresponds to determining the cluster center that maximizes the multiple instance cluster rank. Since all of the data proportions to compute the rank come straight from the clustering results, no additional computation complexity is needed to determine the cluster rank for each cluster. This is an indexed matrix lookup and therefore constant time, and thus not included in the second term of the computation complexity.

\subsubsection{Multiple Instance Cluster Regression (MI-ClusterRegress)}
The MI-ClusterRegress algorithm\cite{wagstaff2008multiple} clusters all of the data regardless of bag structure into $K$ clusters using a Gaussian Mixture Model (GMM)\cite{murphy_2013}. Then, an exemplar point is created in each positive bag for each of the $K$ distributions. An exemplar point is a weighted average of the instances in a bag, where the weights correspond to the membership of that instance to the corresponding cluster. Namely, the exemplar point for cluster $k$ within bag $j$, denoted as $\mathbf{\hat{B^{k}_{j}}}$, is the average of all data points, $\mathbf{x_{i}}$, in bag $j$ weighted by their memberships in cluster $k$, denoted by $R_{i}$. Then, using the exemplars, a regression model is fit to each cluster $k$, using the exemplar points from each bag that correspond to cluster $k$. The eq.s \eqref{MIClustRegressRel1} - \eqref{MIClustRegressRel3} demonstrate how to compute the membership relevance, $R_{i}$, for each instance $\mathbf{x_{i}}$. In the original algorithm, these exemplar points are then used to train $K$ separate regression models to allow each distribution to have their own local regression model.

\begin{equation}
    r_{i} = \text{P}(\mathbf{x_{i}} \in c_{k} | \mathbf{B_{j}}; \boldsymbol{\theta_{c_{k}}}), \forall i
    \label{MIClustRegressRel1}
\end{equation}

\begin{equation}
    z := \sum_{i=1}^{N_{j}^{+}} r_{i}
    \label{MIClustRegressRel2}
\end{equation}

\begin{equation}
    R_{i}:= \frac{r_{i}}{z}, \forall i
    \label{MIClustRegressRel3}
\end{equation}

Here, $\mathbf{x_{i}} \in c_{k} $ means that instance $\mathbf{x_{i}}$ was generated by cluster $c_{k}$. This probability in equation \eqref{MIClustRegressRel1} can be computed using the learned parameters, $\boldsymbol{\theta_{k}}$, from each of the Gaussian Mixture Model distributions. Then a normalization term $z$ is computed for the bag, $\mathbf{B_{j}}$, as the sum of all memberships from bag, $\mathbf{B_{j}}$. Here, $N_{j}^{+}$ is the number of instances in bag $j$. Lastly, each instance's membership is normalized by $z$ to form the relevance, $R_{i}$, of an instance belonging to the $k^{th}$ distribution in bag $j$.

This algorithm has been incorporated into this initialization method, referred to as MI-CR. In this initialization method, the clustering portion of MI-ClusterRegress, as well as creating the exemplar points, is used to reduce the number of instances that must be searched to initialize a target concept. Additionally, the created exemplar points are a combination of the instances in a positive bag, so the initialized exemplar point may be better at representing the variations in target representatives rather than using a single instance as the initialized target concept.

The computation complexity for MI-CR is $O(Ki(N^{+}+N^{-}) + KN^{B^{+}}(N^{+} + N^{-}))$, where $K$ is the number of clusters, $N^{+}$ and $N^{-}$ are the number of positive and negative instances, respectively, $N^{B^{+}}$ is the number of positive bags, and $i$ is the number of iterations until Expectation Maximization for GMM converges. The first term corresponds to GMM's complexity. The second term corresponds to how many exemplar points are being considered as a potential target signature. For each bag, there are $K$ exemplar points generated, so a total of $KN^{B^{+}}$ exemplar points are generated. Then the objective value is computed with complexity of $(N^{+}+N^{-})$ for each of the exemplar points. This will dominate the computation time of MI-CR if there are many bags or many instances, but the computation time can also be largely affected by the stopping condition threshold used for the GMM.

\begin{center}
 \begin{tabular}{|c | c|} 
 \hline
 \multicolumn{2}{||c||}{\textbf{MI-ACE Initialization Methods Time Complexity}} \\ [.5ex] 
  \hline\hline
 Method & Time Complexity \\ [1ex] 
  \hline
 Original &  $O(N^{+}(N^{+}+N^{-}))$\\ [.5ex]
  \hline
 K-Means &  $O(Ki(N^{+}+N^{-}) + K(N^{+} + N^{-}))$\\ [.5ex]
  \hline
 Ranked K-Means & $O(Ki(N^{+}+N^{-}) + K)$ \\ [.5ex]
  \hline
 MI-CR & $O(Ki(N^{+}+N^{-}) + KN^{B^{+}}(N^{+} + N^{-}))$ \\ [.5ex] 
 \hline
\end{tabular}
\end{center}

\section{EXPERIMENTAL RESULTS}

A data collection using handheld electromagnetic induction (EMI) sensors was used to test the various initialization approaches. This data collection consisted of various explosive hazards, and for testing purposes was broken into three groups. The three groups were all high metal targets, all low metal targets, and all targets including no metal. Additionally, two different EMI sensors were used, sensor A and sensor B. Experiments were run with both the initialized signature and the optimized signature. With this configuration, a total of 12 experiments were run, six for the initialized signature, and six with the optimized signature.

\begin{center}
 \begin{tabular}{|c | c|} 
 \hline
 \multicolumn{2}{||c||}{\textbf{Number of Targets in Data Subsets}} \\ [.5ex] 
 \hline\hline
  Data Subset & Number of Targets \\ [1ex] 
 \hline
 Sensor A (Metal) & 39 \\ 
 \hline
 Sensor A (Low) & 70 \\ 
 \hline
 Sensor A (all) & 167 \\ 
  \hline
 Sensor B (Metal) & 9 \\ 
  \hline
 Sensor B (Low) & 20 \\ 
  \hline
 Sensor B (all) & 39 \\ [.5ex] 
 \hline
\end{tabular}
\end{center}

The algorithm was trained using lane based cross validation. For example, if a site consisted of five lanes, during one fold of cross validation, four lanes would be used for training and the other lane would be used for test. Each lane consisted of an unknown multiple of grids, where each grid contained a single explosive hazard. Each positive bag was generated from a single grid. The samples inside a predicted response radius of the target's spatial center were taken to be the samples in the positive bag. A single negative bag was generated using all samples from the low and no metal grids blank sweeps. The blank sweeps were created at the beginning portion of a grid. It was seen that high metal grids would often have target response bleed over in to the blank sweeps, so only the low and no metal blank sweeps were used for the negative bag.

For, only the initialized signatures were used to generate confidence maps on the test data. Namely, no optimization was done after the learned target signature was initialized. Then, optimization was included with each of the techniques. For all of the proposed techniques, five clusters were used. Once a signature was learned, the ACE similarity statistic was used to generate confidence maps, a confidence for every sample along a test sweep. The generated confidences, along with the corresponding spatial coordinates, were then passed into a mean shift clustering algorithm\cite{Fukunaga1975} to generate alarms. Mean shift clustering was used to group the spatial regions of the sweep that had similar confidences into separate clusters. Each cluster's center was used as the cluster's alarm location.

In practice, a larger uniform response is desired for detecting explosive hazards. The operator would be able to determine if an explosive hazard exists earlier, and be able to determine its' shape and location with more accuracy. This was taken into account for setting the score of a generated alarm. The samples that belong to a generated alarm are those in the initial mean shift cluster and those within an allowable distance, $d_{halo}$, of the cluster center, $c_{center}$. The allowable distance was set to 0.25 meters for these experiments. The score of an alarm was computed as the weighted average of the confidences that belong to that alarm, shown in equation \eqref{AlarmScore}.

\begin{equation}
	Score = \frac{1}{N_{c}} \sum_{i = 1}^{N_c} (w_{i}*conf_{i})
	\label{AlarmScore}
\end{equation}

Where $N_{c}$ is the number of samples in the alarm, and $conf_{i}$ is the confidence associated with the $i^{th}$ sample, $x_{i}$. The weight, $w_{i}$, corresponding to the $i^{th}$ sample, is the spatial Euclidean distance of the sample to the center of the cluster, $c_{center}$, shown in equation \eqref{AlarmScoreDist}. The weight is divided by the maximum allowable distance, $d_{halo}$, to normalize the weight from 0 to 1. 

\begin{equation}
	w_{i} = \frac{\vert \vert x_{i} - c_{center} \vert \vert}{d_{halo}}
	\label{AlarmScoreDist}
\end{equation}

This was done to boost the score of alarms that had a larger surface area, as this is desired in practice. The label of the alarm was determined by whether the cluster center fell into the expected response radius of an explosive hazard. Finally, these alarms and labels were used to generate ROC curves for the different experiments.

\begin{center}
	\begin{figure}[H]
	\centering
	\begin{subfigure}{.33\textwidth}
  		\centering
        \includegraphics[width=\linewidth]{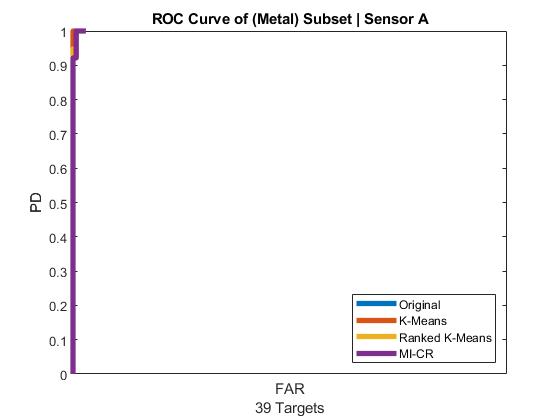}
	\end{subfigure}%
	\begin{subfigure}{.33\textwidth}
  		\centering
        \includegraphics[width=\linewidth]{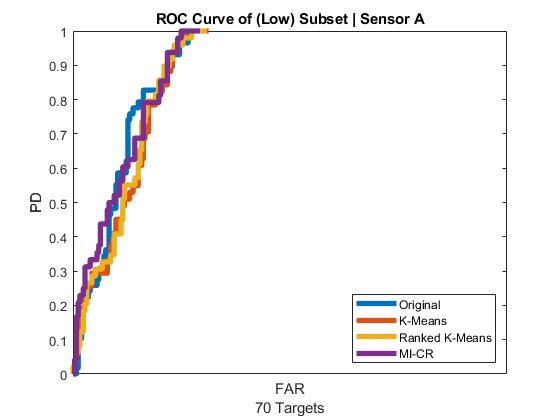}
	\end{subfigure}
	\begin{subfigure}{.33\textwidth}
  		\centering
        \includegraphics[width=\linewidth]{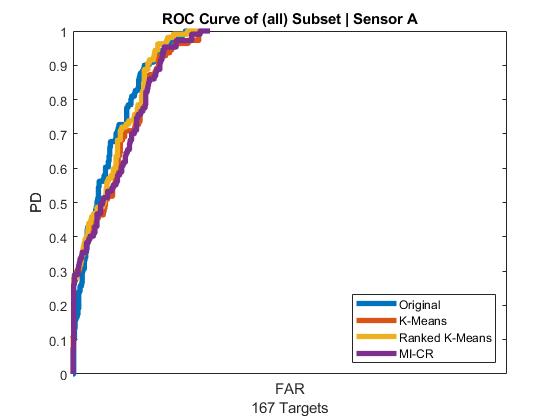}
	\end{subfigure}
	\caption{MI-ACE ROC curves using the initialized signature from the original initialization method and the three proposed initialization methods using Sensor A for high metal, low metal, and all explosive hazards, from left to right.}
	\label{SensorAInitROCs}
	\end{figure}
\end{center}

\begin{center}
	\begin{figure}[H]
	\centering
	\begin{subfigure}{.33\textwidth}
  		\centering
        \includegraphics[width=\linewidth]{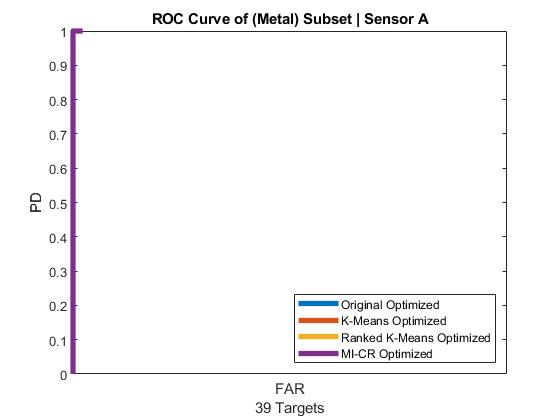}
	\end{subfigure}%
	\begin{subfigure}{.33\textwidth}
  		\centering
        \includegraphics[width=\linewidth]{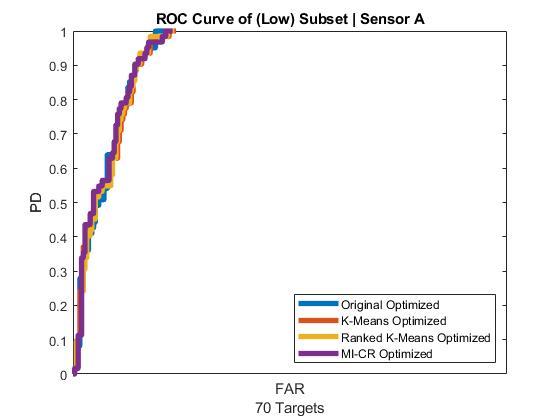}
	\end{subfigure}
	\begin{subfigure}{.33\textwidth}
  		\centering
        \includegraphics[width=\linewidth]{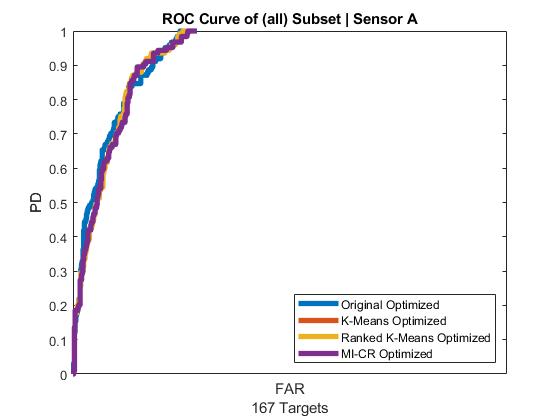}
	\end{subfigure}
	\caption{MI-ACE ROC curves using the corresponding optimized signature from the original initialization method and the three proposed initialization methods using Sensor A for high metal, low metal, and all explosive hazards, from left to right.}
	\label{SensorAOptROCs}
	\end{figure}
\end{center}

\begin{center}
	\begin{figure}[H]
	\centering
	\begin{subfigure}{.33\textwidth}
  		\centering
        \includegraphics[width=\linewidth]{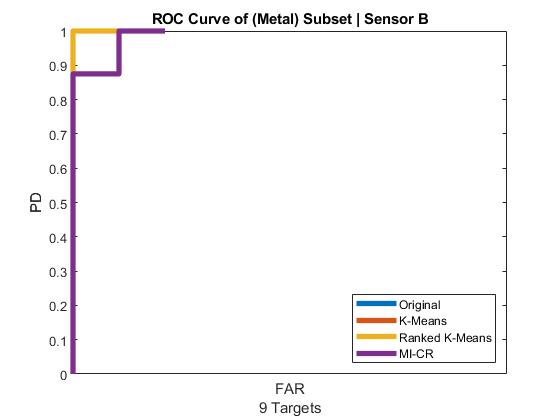}
	\end{subfigure}%
	\begin{subfigure}{.33\textwidth}
  		\centering
        \includegraphics[width=\linewidth]{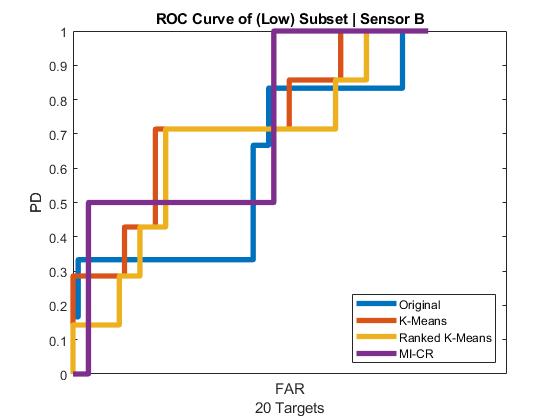}
	\end{subfigure}
	\begin{subfigure}{.33\textwidth}
  		\centering
        \includegraphics[width=\linewidth]{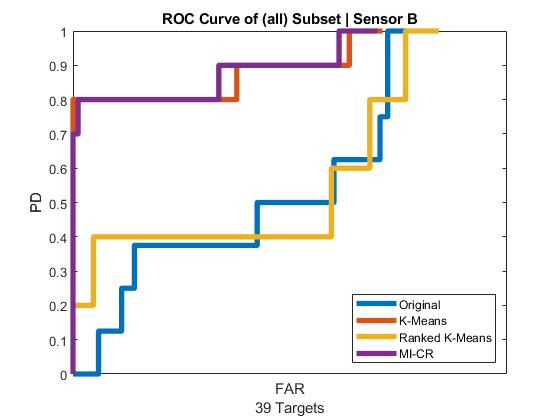}
	\end{subfigure}
	\caption{MI-ACE ROC curves using the initialized signature from the original initialization method and the three proposed initialization methods using Sensor A for high metal, low metal, and all explosive hazards, from left to right.}
	\label{SensorBInitROCs}
	\end{figure}
\end{center}

\begin{center}
	\begin{figure}[H]
	\centering
	\begin{subfigure}{.33\textwidth}
  		\centering
        \includegraphics[width=\linewidth]{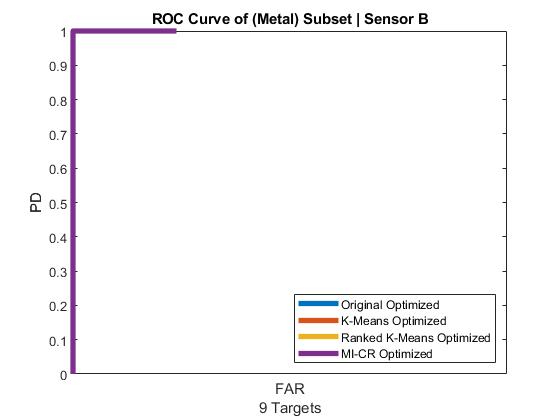}
	\end{subfigure}%
	\begin{subfigure}{.33\textwidth}
  		\centering
        \includegraphics[width=\linewidth]{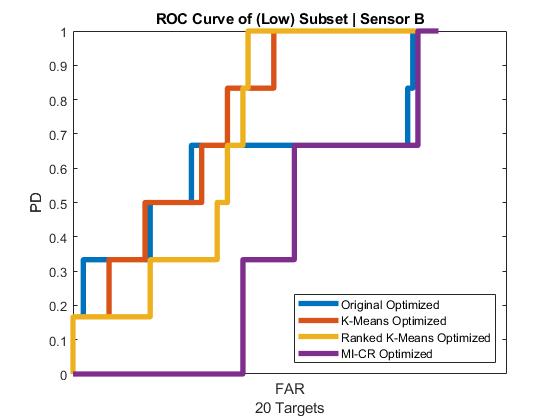}
	\end{subfigure}
	\begin{subfigure}{.33\textwidth}
  		\centering
        \includegraphics[width=\linewidth]{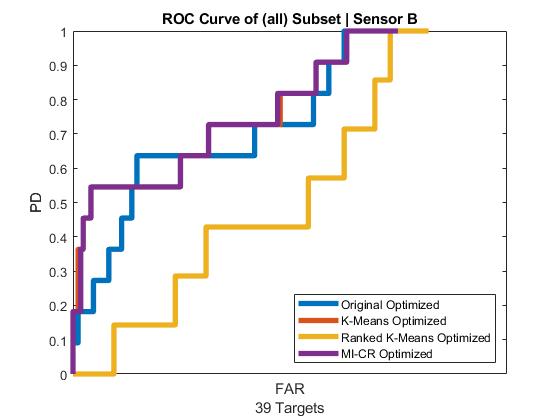}
	\end{subfigure}
	\caption{MI-ACE ROC curves using the corresponding optimized signature from the original initialization method and the three proposed initialization methods using Sensor B for high metal, low metal, and all explosive hazards, from left to right.}
	\label{SensorBOptROCs}
	\end{figure}
\end{center}

\begin{center}
 \begin{tabular}{|c | c c c c|}
 \hline
 \multicolumn{5}{||c||}{\textbf{Initialization Run Time Comparison [ ms ]}} \\ [1ex] 
 \hline\hline
 Experiment & Original & K-Means & Ranked K-Means & MI-CR \\ [.5ex] 
 \hline
 Sensor A (Metal) & 816.0 & 6.9 & 10.5 & 37.1 \\ 
 \hline
 Sensor A (Low) & 2,216.7 & 28.6 & 18.2 & 98.3 \\ 
 \hline
 Sensor A (all) & 40,431.4 & 65.1 & 60.0 & 250.0 \\ 
  \hline
 Sensor B (Metal) & 42.2 & 1.0 & 1.4 & 9.4 \\ 
  \hline
 Sensor B (Low) & 211.5 & 1.7 & 4.7 & 81.7 \\ 
  \hline
 Sensor B (all) & 678.5 & 15.4 & 7.4 & 64.6 \\ [.5ex] 
 \hline
 \textbf{Average} & \textbf{7,399.4} & \textbf{19.8} & \textbf{17.0} & \textbf{90.2} \\ 
 \hline
\end{tabular}
\end{center}

\section{DISCUSSION AND FUTURE WORK}

\begin{center}
	\begin{figure}[H]
	\centering
	\begin{subfigure}{.4\textwidth}
  		\centering
        \includegraphics[width=\linewidth]{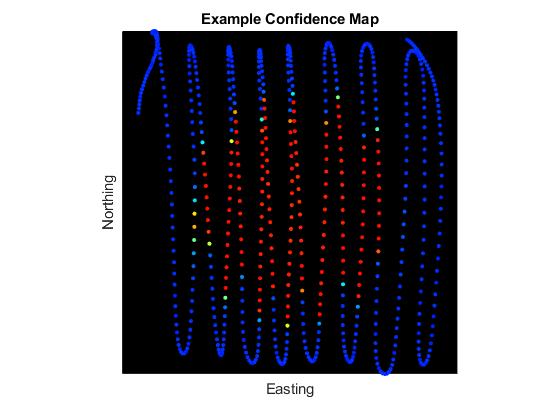}
	\end{subfigure}%
	\begin{subfigure}{.4\textwidth}
  		\centering
        \includegraphics[width=\linewidth]{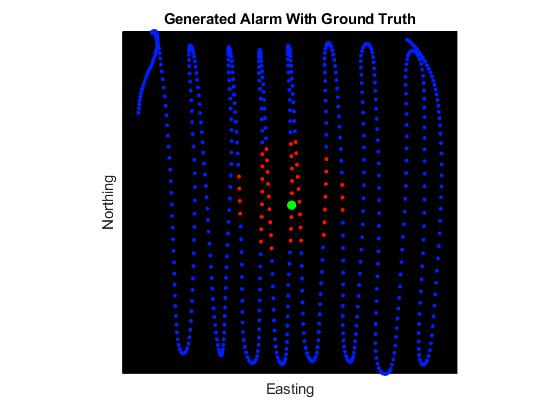}
	\end{subfigure}
	\caption{An example confidence map (left) with it's corresponding alarm generation plotted over the ground truth (right). The Alarm generation works as expected.}
	\label{GoodMeanShift}
	\end{figure}
\end{center}

\begin{center}
	\begin{figure}[H]
	\centering
	\begin{subfigure}{.4\textwidth}
  		\centering
        \includegraphics[width=\linewidth]{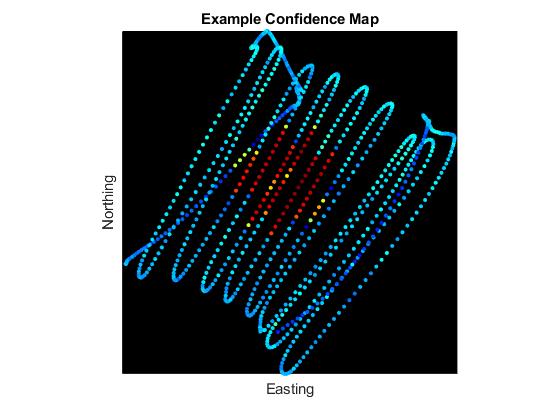}
	\end{subfigure}%
	\begin{subfigure}{.4\textwidth}
  		\centering
        \includegraphics[width=\linewidth]{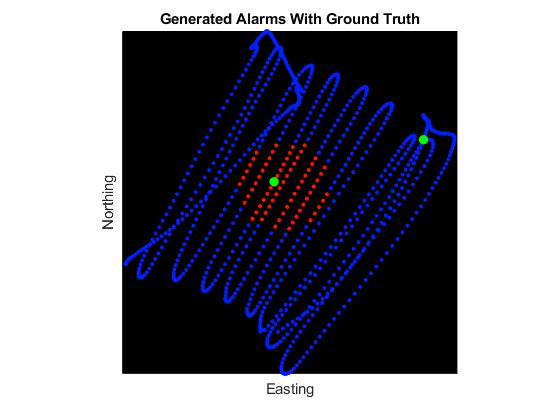}
	\end{subfigure}
	\caption{An example confidence map (left) with it's corresponding alarm generation plotted over the ground truth (right). Two Alarms are generated even though it appears clear there should be only one generated.}
	\label{BadMeanShift1}
	\end{figure}
\end{center}

\begin{center}
	\begin{figure}[H]
	\centering
	\begin{subfigure}{.4\textwidth}
  		\centering
        \includegraphics[width=\linewidth]{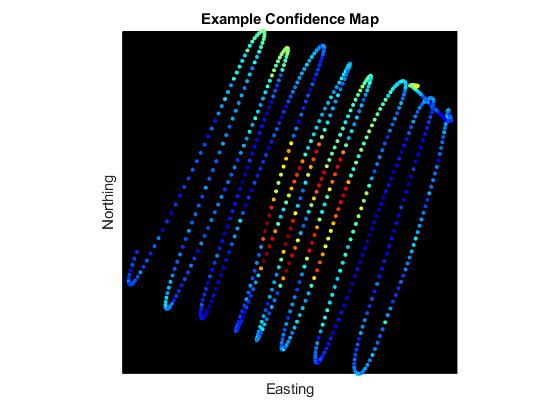}
	\end{subfigure}%
	\begin{subfigure}{.4\textwidth}
  		\centering
        \includegraphics[width=\linewidth]{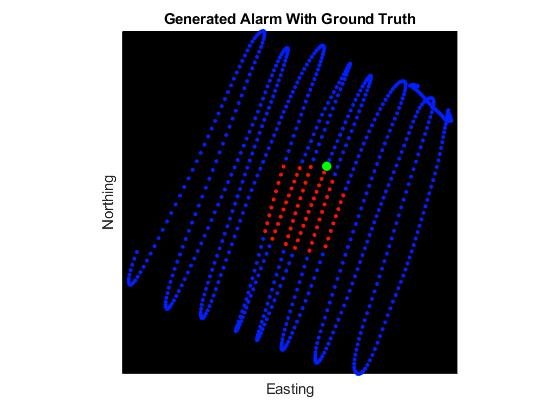}
	\end{subfigure}
	\caption{An example confidence map (left) with it's corresponding alarm generation plotted over the ground truth (right). The generated alarm falls just outside the ground truth and is considered as a false alarm.}
	\label{BadMeanShift2}
	\end{figure}
\end{center}

The focus of this paper is to analyze the various initialization approaches proposed. In most cases the ROC curve generation process works as expected. It can be seen in figure \ref{GoodMeanShift} that an alarm is generated in the center of the high confidence region as expected. The alarm generation is not perfect though, and has it's flaws. The problem with the alarm generation is twofold, first the mean shift clustering does not always perform as expected. For example taking the downtrack confidence map and generating it's alarms, shown in figure \ref{BadMeanShift1}, we can see that mean shift generates multiple alarms. It appears clear that there should only be one alarm, but mean shift generates an additional alarm. Second, the response radius of an explosive hazard is unknown and changes based on preprocessing and the algorithm being used. So a generated alarm that should be truly considered a true target alarm, can fall just outside what was determined to be the ground truth and cause a false alarm. An example of this is shown in figure \ref{BadMeanShift2}, where the alarm generated came from the response of the explosive hazard, but due to an unknown true response radius, the alarm is considered a false alarm.

With these examples in place, this promotes a need for creating a variable ground truth radius. The radius could be based on explosive hazard type or even algorithm, as it has been noticed that the response radius changes with different algorithms. If the ground truth can be variable, the ROC generation curves would be more accurate, and a more definitive analysis can be done.

With this said, the MI-ACE initialization performances, with and without optimization are compared. It can be seen that the different initialization approaches perform similarly for the majority of the experiments. All of the approaches are able to detect the high metal targets, and when the learned signature is optimized, all of the techniques find the high metal targets with no false alarms. In the experiments of sensor B, there are folds of the cross validation where the testing set has target types that do not exist in the training set. This is why it is believed that the initialized signature ROC with the all subset of sensor B, shows K-Means and MI-CR performing better than the original and Ranked K-Means. It is believed that K-Means and MI-CR are able to initialize a signature that generalizes the target class better. Whereas the original initialization technique chooses a single sample that maximizes the objective function for the training set data, which may not generalize well to the target class as a whole or the target types in the test set. Additionally, it is expected that when there are more targets available for training, that the initialization techniques would perform more similarly like they do with sensor A which has approximately four times as many training examples as sensor B.

Furthermore, it can be seen that if the training and test data contain the same target types, the case of sensor A, any of these initialization methods will provide marginally no difference in performance when optimization is performed. This means that as long as optimization is done, the original costly initialization can be replaced with one of these variations to save run time. This is largely the case because of the nature of optimization. If two initialized signatures are similar, the optimization will likely optimize the signatures to be even more similar. This is because of how signatures are optimized. The optimized signature is the average of the positive bag representations, minus the average of the background. So often times, different initialized signatures will select the same or similar positive bag representatives and the updated signature will start averaging towards the same result, causing the resulting optimized signatures to be very similar in nature. This was noticed by analyzing how signatures change in optimization and can be confirmed by comparing Figures \ref{SensorAInitROCs} and \ref{SensorAOptROCs}. It can be seen that the difference in ROC curves is smaller for the optimized signatures than the initialized signatures. This is likely because the optimized signatures are more similar than the initialized signatures.

The run time analysis shows that all of the proposed techniques are faster than the original, and in the case of the K-Means clustering techniques, on average they are faster on the order of 100. Furthermore, this is consistent with the time complexity analysis done in section \ref{SectionApproaches} and shows that with a decrease in run time, the same performance can be obtained with any of the proposed initialization techniques.

\section{Acknowledgments}

This work was funded by Army Research Office grant number W911NF-17-1-0213 to support the US Army RDECOM CERDEC NVESD. The views and conclusions contained in this document
are those of the authors and should not be interpreted as representing the official policies either expressed or implied, of the Army Research Office, Army Research Laboratory, or the U.S. Government. The U.S. Government is authorized to reproduce and distribute reprints for Government purposes notwithstanding any copyright notation hereon.


\bibliography{Bibliography}   

\begin{thebibliography}{10}

\bibitem{zare2015multiple}
Zare, A., Cook, M., Alvey, B., and Ho, D.~K., ``Multiple instance dictionary
  learning for subsurface object detection using handheld emi,'' in [{\em
  Detection and Sensing of Mines, Explosive Objects, and Obscured Targets
  XX}{\nolinebreak\hspace{0.1em}]},   {\bf 9454},  94540G, International
  Society for Optics and Photonics (2015).

\bibitem{karem2015fuzzy}
Karem, A. and Frigui, H., ``Fuzzy clustering of multiple instance data,'' in
  [{\em Fuzzy Systems (FUZZ-IEEE), 2015 IEEE International Conference
  on}{\nolinebreak\hspace{0.1em}]},   1--7, IEEE (2015).

\bibitem{alvey2017fourier}
Alvey, B., Ho, D.~K., and Zare, A., ``Fourier features for explosive hazard
  detection using a wideband electromagnetic induction sensor,'' in [{\em
  Detection and Sensing of Mines, Explosive Objects, and Obscured Targets
  XXII}{\nolinebreak\hspace{0.1em}]},   {\bf 10182},  101820E, International
  Society for Optics and Photonics (2017).

\bibitem{cook2016buried}
Cook, M., Zare, A., and Ho, D.~K., ``Buried object detection using handheld
  wemi with task-driven extended functions of multiple instances,'' in [{\em
  Detection and Sensing of Mines, Explosive Objects, and Obscured Targets
  XXI}{\nolinebreak\hspace{0.1em}]},   {\bf 9823},  98230A, International
  Society for Optics and Photonics (2016).

\bibitem{dietterich1997solving}
Dietterich, T.~G., Lathrop, R.~H., and Lozano-P{\'e}rez, T., ``Solving the
  multiple instance problem with axis-parallel rectangles,'' {\em Artificial
  intelligence}~{\bf 89}(1-2),  31--71 (1997).

\bibitem{wei2011landmine}
Wei, M.-H., Scott, W.~R., and McClellan, J.~H., ``Landmine detection using the
  discrete spectrum of relaxation frequencies,'' in [{\em 2011 IEEE
  International Geoscience and Remote Sensing
  Symposium}{\nolinebreak\hspace{0.1em}]},   834--837, IEEE (2011).

\bibitem{mayer2003object}
Mayer, R., Bucholtz, F., and Scribner, D., ``Object detection by using"
  whitening/dewhitening" to transform target signatures in multitemporal
  hyperspectral and multispectral imagery,'' {\em IEEE transactions on
  geoscience and remote sensing}~{\bf 41}(5),  1136--1142 (2003).

\bibitem{kraut2001adaptive}
Kraut, S., Scharf, L.~L., and McWhorter, L.~T., ``Adaptive subspace
  detectors,'' {\em IEEE Transactions on signal processing}~{\bf 49}(1),  1--16
  (2001).

\bibitem{wu2016two}
Wu, J.-C. and Wu, K.-B., ``Two-stage process for improving the performance of
  hyperspectral target detection,'' in [{\em 2016 8th Workshop on Hyperspectral
  Image and Signal Processing: Evolution in Remote Sensing
  (WHISPERS)}{\nolinebreak\hspace{0.1em}]},   1--4, IEEE (2016).

\bibitem{zare2018discriminative}
Zare, A., Jiao, C., and Glenn, T., ``Discriminative multiple instance
  hyperspectral target characterization,'' {\em IEEE transactions on pattern
  analysis and machine intelligence}~{\bf 40}(10),  2342--2354 (2018).

\bibitem{macqueen1967some}
MacQueen, J. et~al., ``Some methods for classification and analysis of
  multivariate observations,'' in [{\em Proceedings of the fifth Berkeley
  symposium on mathematical statistics and
  probability}{\nolinebreak\hspace{0.1em}]},   {\bf 1}(14),  281--297, Oakland,
  CA, USA (1967).

\bibitem{wagstaff2008multiple}
Wagstaff, K.~L., Lane, T., and Roper, A., ``Multiple-instance regression with
  structured data,'' in [{\em Data Mining Workshops, 2008. ICDMW'08. IEEE
  International Conference on}{\nolinebreak\hspace{0.1em}]},   291--300, IEEE
  (2008).

\bibitem{murphy_2013}
Murphy, K.~P.,  [{\em Machine learning: a probabilistic
  perspective}{\nolinebreak\hspace{0.1em}]}, Cram101 (2013).

\bibitem{Fukunaga1975}
Fukunaga, K. and Hostetler, L., ``The estimation of the gradient of a density
  function, with applications in pattern recognition,'' {\em IEEE Transactions
  on Information Theory}~{\bf 21},  32--40 (January 1975).

\end{thebibliography}
\bibliographystyle{spiebib}   

\end{document}